\documentclass[11pt, a4paper]{article}

\usepackage[utf8]{inputenc} 
\usepackage[T1]{fontenc}    
\usepackage{geometry}       
\usepackage{microtype}      
\usepackage{xcolor}         
\usepackage{hyperref}       
\usepackage{enumitem}       
\usepackage{mathpazo}       
\usepackage{titlesec}       
\usepackage{amsmath, amsfonts, amssymb} 
\usepackage{booktabs}       

\usepackage{tikz}
\usetikzlibrary{positioning, arrows.meta, calc, matrix, backgrounds}
\usepackage[most]{tcolorbox}
\tcbuselibrary{listings, breakable} 

\usepackage[font=small,labelfont=bf]{caption}

\definecolor{darkblue}{RGB}{0, 43, 91}

\hypersetup{
    colorlinks=true,
    linkcolor=darkblue,
    citecolor=darkblue,
    urlcolor=darkblue,
}

\renewenvironment{abstract}
 {\small
  \begin{center}
  \bfseries \abstractname\vspace{-.5em}\vspace{0pt}
  \end{center}
  \list{}{%
    \setlength{\leftmargin}{0.75cm}%
    \setlength{\rightmargin}{\leftmargin}%
  }%
  \item\relax}
 {\endlist}

\newtcolorbox{examplebox}[1]{
  colback=darkblue!4,
  colframe=darkblue!80,
  fonttitle=\bfseries\small,
  title=#1,
  arc=2pt,
  boxrule=0.7pt,
  left=5pt, right=5pt, top=4pt, bottom=4pt
}

\title{\vspace{-2em}\color{darkblue}\Large\bfseries Teaching LLMs String Matching, Backtracking, and Error Recovery to Deduce Bases and Truth Tables for the Combinatorially Exploding Bit Manipulation Puzzles}

\author{
  \large \textbf{Prateek Agnihotri} \quad \textbf{Sanchit Jain} \quad \textbf{Prabhat Agnihotri} \\[0.2em] 
  \large \textbf{Aditya Prasad} \quad \textbf{Shubham Jain} \\[0.6em]
  \small \texttt{\{prateekkumaragnihotri, sanchitsj4321, agnihotriprabhat1999\}@gmail.com} \\
  \small \texttt{\{prasadaditya97, shubham2605jain\}@gmail.com} \\[0.6em]
  \normalsize \textbf{NVIDIA Nemotron Challenge (Kaggle) - Winners of} \\
  \normalsize \textbf{Best Fine-tuning Method \& Best Data/Synthetic Data Method}
}

\date{\vspace{-1.2cm}} 

\begin{document}

\maketitle

{
\renewcommand{\thefootnote}{}
\footnotetext{The complete Python source code for our deterministic Set Cover solver and synthetic Chain-of-Thought (CoT) data generation pipeline is publicly available at: \url{https://www.kaggle.com/code/prateekagnihotri/bit-manipulation-cot-bit-flip-s-original}}
}

\begin{abstract}
This paper presents the key ideas and algorithmic innovations developed during our participation in the NVIDIA Nemotron Model Reasoning Challenge, focusing specifically on the Bit Manipulation Puzzles, which was widely considered one of the most challenging tasks to fully solve. In this task, the objective is to discover a hidden logical rule that transforms a given set of input binary strings into outputs, and then accurately apply that rule to a new, unseen input. Large Language Models (LLMs) notoriously struggle with this; traditional methods force them to simulate complex boolean logic and arithmetic in their heads, which frequently leads to hallucinations. Furthermore, the search space of possible bitwise operations—comprising arbitrary combinations of shifts, rotations, and logic gates—suffers from a severe combinatorial explosion. Because this makes traditional logic-gate deduction computationally intractable for LLMs, we present a novel approach that abandons arithmetic logic entirely in favor of string similarity, structured search, and autonomous error recovery.

\vspace{0.5em}\noindent
Our core contributions are:
\begin{itemize}[leftmargin=1.5em, itemsep=0.3em, topsep=0.3em]
    \item \textbf{Bases and Truth Table Formulation:} We reframe logic-gate deduction into a base-selection task, leveraging string similarity (minimal bit flips) between outputs to isolate primitive transformations (``bases'') and deduce their truth tables without complex arithmetic.
    \item \textbf{Backtracking DFS and Error Recovery:} We formalize a search process that tests candidate bases, detects logical collisions across examples in a puzzle, and backtracks upon failure to perform robust error recovery.
    \item \textbf{Bit Tokenization and Interactive Reasoning SFT:} We force the tokenizer to encode binary strings as individual, single-bit tokens, and use dynamic masking to simulate external oracle feedback—training the model to hypothesize, self-evaluate, and backtrack natively.
\end{itemize}

\vspace{0.5em}\noindent
Evaluated on bit manipulation puzzles, our approach achieved $> \! 96\%$ validation accuracy—representing the highest performance in this category by any participating team. These algorithmic and training innovations ultimately earned our team the \textbf{Best Fine-tuning Method} and \textbf{Best Data/Synthetic Data Method} awards in the contest, alongside a 7th Place overall finish on leaderboard.
\end{abstract}

\vspace{0.3cm}
\section{Introduction}

Modern Artificial Intelligence excels at generating fluent text and heuristic reasoning, but exact, deterministic algorithmic logic remains one of its most difficult frontiers. A rigorous benchmark for this capability is the ``Bit Manipulation'' puzzle, featured as a core category in the NVIDIA Nemotron Model Reasoning Challenge. In these puzzles, an AI system must reverse-engineer a hidden mathematical formula simply by observing a few examples of how one 8-bit binary string transforms into another. To solve the puzzle, the model must deduce the exact sequence of operations applied to the inputs and accurately predict the output for a novel, unseen binary string.

\begin{examplebox}{Example Problem: The Bit Manipulation Puzzle}
\small
In Alice's Wonderland, a secret bit manipulation rule transforms 8-bit binary numbers using operations like bit shifts, rotations, XOR, AND, OR, NOT, and majority or choice functions. Deduce the underlying rule to predict the target output.
\begin{itemize}[leftmargin=1.5em, itemsep=0pt, topsep=2pt, parsep=0pt]
    \item \textbf{Example 1:} \texttt{10100011} $\longrightarrow$ \texttt{11011001}
    \item \textbf{Example 2:} \texttt{01100110} $\longrightarrow$ \texttt{10001101}
    \item \textbf{Example 3:} \texttt{11110110} $\longrightarrow$ \texttt{11011011}
    \item \textbf{Target:} \quad \,\, \texttt{01001010} $\longrightarrow$ \textbf{\texttt{????????}}
\end{itemize}
\end{examplebox}

Traditional methodologies approach this sequence-to-sequence translation by prompting the Large Language Model (LLM) to reverse-engineer an abstract syntax tree of boolean logic gates (e.g., \texttt{AND}, \texttt{OR}, \texttt{XOR}) and spatial operators (e.g., left shifts, right shifts, circular rotations). However, this approach collides directly with the mathematical reality of the problem: the search space of possible operations is astronomically large. 

To illustrate this intractability, consider the full feature space of 22 possible base transformations (the original bit \texttt{x}, plus 7 left shifts, 7 right shifts, and 7 circular rotations). If a hidden rule relies on exactly three of these bases, there are $\binom{22}{3} = 1,540$ unique base combinations. However, to form an executable mathematical rule, these three bases must be ordered into a syntax tree and connected using the allowed operations (\texttt{XOR}, \texttt{AND}, \texttt{OR}, \texttt{NOT}, as well as \texttt{majority} or \texttt{choice} functions). For any three selected bases, there are $3! = 6$ ways to order them. If we connect them using two sequential binary gates (including their negated variants, yielding 6 options per gate), we generate $3! \times 6 \times 6 = 216$ distinct structural equations per subset. Thus, even for a rudimentary three-variable rule, an LLM attempting to brute-force the formula would have to evaluate a search space exceeding $330,000$ unique combinations ($1,540 \times 216$).

Because this search space expands exponentially with each additional logical step, standard heuristic guessing or structured arithmetic search quickly becomes intractable. Auto-Regressive architectures lack the internal working memory to mentally simulate shifted binary arrays and evaluate complex boolean algebra simultaneously across thousands of theoretical paths. Consequently, forcing LLMs down this arithmetic deduction route invariably results in severe hallucinations and logical dead-ends.

To overcome this fundamental computational barrier, we present a paradigm shift in how bitwise reasoning is handled. Rather than forcing the model to perform arithmetic deduction, we reformulate the entire puzzle into a discrete feature-selection and string-matching problem. We decompose the input space into a finite set of spatial pointers (``Bases'') and leverage string similarity metrics—specifically, tracking minimal bitflips between varying outputs—to directly isolate the exact variables driving the state changes. 

Furthermore, we recognize that a robust reasoning system must possess the ability to hypothesize, verify against an external ground truth, and recover from logical collisions. To instill this System-2 capability, we introduce a novel Supervised Fine-Tuning (SFT) methodology: \textbf{Interactive Reasoning SFT via Dynamic Masking}. By coupling strict single-bit tokenization with masked environmental feedback, we successfully train the LLM to act as an autonomous agent that navigates a Depth-First Search (DFS) tree, deduces empirical truth tables, and backtracks natively when its logic fails.

\vspace{0.2cm}
\noindent \textbf{Paper Organization:}
The remainder of this paper is organized as follows: 
\textbf{Section 2} establishes our core conceptual framework, detailing the deconstruction of bit sequences into spatial bases, empirical truth tables, and the isolation of output-influencing bases via minimal bitflips. 
\textbf{Section 3} formalizes our deterministic solver, showcasing our backtracking Depth-First Search (DFS) and global collision verification. 
\textbf{Section 4} presents our machine learning methodologies, including token-alignment strategies and interactive reasoning via dynamic masking. 
Finally, \textbf{Section 5} evaluates our empirical results and details our error analysis under strict hardware constraints.

\section{Preliminaries and Core Concepts}

To understand why our approach abandons traditional logic-gate simulation, it is first necessary to establish the conceptual framework we use to analyze bit manipulation. Instead of viewing the problem through the lens of complex logic gates and bitwise arithmetic, we transformed the sequence-to-sequence translation into a discrete base-selection and string-matching problem. This requires understanding three foundational concepts: Bases, Empirical Truth Tables, and Isolating Output-Influencing Bases via Minimal Bitflips.

\subsection{Bases}

\begin{figure}[h]
\centering
\begin{tikzpicture}[
    scale=0.82, transform shape, 
    bit/.style={draw, minimum size=0.45cm, font=\ttfamily\scriptsize, fill=gray!5, minimum width=0.45cm, inner sep=0pt},
    targetbit/.style={draw=red!80!black, thick, minimum size=0.45cm, font=\ttfamily\scriptsize\bfseries, fill=red!10, inner sep=0pt},
    phantomL/.style={draw=teal!80!black, dashed, thick, minimum size=0.45cm, font=\ttfamily\scriptsize, fill=teal!5, text=teal!80!black, inner sep=0pt},
    phantomR/.style={draw=blue!80!black, dashed, thick, minimum size=0.45cm, font=\ttfamily\scriptsize, fill=blue!5, text=blue!80!black, inner sep=0pt},
    arrow/.style={-{Latex[scale=0.7]}, thick},
    lbl/.style={fill=white, inner sep=1.5pt, font=\ttfamily\tiny\bfseries}
]

    \node[font=\Large\bfseries] (arr_eq) at (7.5, 5.0) {$\longrightarrow$};
    
    \node[bit, left=3.6cm of arr_eq] (eq_in0) {1};
    \node[bit, right=0cm of eq_in0] (eq_in1) {0};
    \node[bit, right=0cm of eq_in1] (eq_in2) {1};
    \node[bit, right=0cm of eq_in2] (eq_in3) {0};
    \node[bit, right=0cm of eq_in3] (eq_in4) {0};
    \node[bit, right=0cm of eq_in4] (eq_in5) {0};
    \node[bit, right=0cm of eq_in5] (eq_in6) {1};
    \node[bit, right=0cm of eq_in6] (eq_in7) {1};

    \node[bit, right=0.5cm of arr_eq] (eq_out0) {1};
    \node[bit, right=0cm of eq_out0] (eq_out1) {1};
    \node[bit, right=0cm of eq_out1] (eq_out2) {0};
    \node[bit, right=0cm of eq_out2] (eq_out3) {1};
    \node[bit, right=0cm of eq_out3] (eq_out4) {1};
    \node[bit, right=0cm of eq_out4] (eq_out5) {0};
    \node[bit, right=0cm of eq_out5] (eq_out6) {0};
    \node[bit, right=0cm of eq_out6] (eq_out7) {1};

    \begin{scope}[xshift=0cm, yshift=0cm]
        \node[font=\bfseries\scriptsize, color=blue!80!black, anchor=west] at (-0.2, 3.8) {1. Right Shifts (R1-R7)};
        
        \node[phantomR] (i1_P1) at (0, 3.0) {0}; 
        
        \node[bit, right=0.15cm of i1_P1] (i1_0) {1};
        \node[bit, right=0cm of i1_0] (i1_1) {0};
        \node[bit, right=0cm of i1_1] (i1_2) {1};
        \node[bit, right=0cm of i1_2] (i1_3) {0};
        \node[bit, right=0cm of i1_3] (i1_4) {0};
        \node[bit, right=0cm of i1_4] (i1_5) {0};
        \node[bit, right=0cm of i1_5] (i1_6) {1};
        \node[bit, right=0cm of i1_6] (i1_7) {1};
        
        \node[bit] (o1_0) at (0.6, 0.0) {1}; 
        \node[bit, right=0cm of o1_0] (o1_1) {1};
        \node[bit, right=0cm of o1_1] (o1_2) {0};
        \node[bit, right=0cm of o1_2] (o1_3) {1};
        \node[bit, right=0cm of o1_3] (o1_4) {1};
        \node[bit, right=0cm of o1_4] (o1_5) {0};
        \node[targetbit, right=0cm of o1_5] (o1_6) {0}; 
        \node[bit, right=0cm of o1_6] (o1_7) {1};
        
        \draw[arrow, color=blue!80!black] (i1_5.south) to[out=-90, in=90] node[pos=0.35, lbl, text=blue!80!black] {R1} (o1_6.north);
        \draw[arrow, color=blue!80!black] (i1_4.south) to[out=-90, in=90] node[pos=0.45, lbl, text=blue!80!black] {R2} (o1_6.north);
        \draw[arrow, color=blue!80!black] (i1_3.south) to[out=-90, in=90] node[pos=0.55, lbl, text=blue!80!black] {R3} (o1_6.north);
        \draw[arrow, color=blue!80!black] (i1_2.south) to[out=-90, in=90] node[pos=0.65, lbl, text=blue!80!black] {R4} (o1_6.north);
        \draw[arrow, color=blue!80!black] (i1_1.south) to[out=-90, in=90] node[pos=0.75, lbl, text=blue!80!black] {R5} (o1_6.north);
        \draw[arrow, color=blue!80!black] (i1_0.south) to[out=-90, in=90] node[pos=0.85, lbl, text=blue!80!black] {R6} (o1_6.north);
        \draw[arrow, color=blue!80!black] (i1_P1.south) to[out=-90, in=90] node[pos=0.92, lbl, text=blue!80!black] {R7} (o1_6.north);
    \end{scope}

    \begin{scope}[xshift=5.2cm, yshift=0cm]
        \node[font=\bfseries\scriptsize, color=purple!80!black, anchor=west] at (-0.2, 3.8) {2. Circular Shifts (C1-C7)};
        \node[bit] (i2_0) at (0, 3.0) {1};
        \node[bit, right=0cm of i2_0] (i2_1) {0};
        \node[bit, right=0cm of i2_1] (i2_2) {1};
        \node[bit, right=0cm of i2_2] (i2_3) {0};
        \node[bit, right=0cm of i2_3] (i2_4) {0};
        \node[bit, right=0cm of i2_4] (i2_5) {0};
        \node[bit, right=0cm of i2_5] (i2_6) {1};
        \node[bit, right=0cm of i2_6] (i2_7) {1};
        
        \node[bit] (o2_0) at (0, 0.0) {1};
        \node[bit, right=0cm of o2_0] (o2_1) {1};
        \node[bit, right=0cm of o2_1] (o2_2) {0};
        \node[bit, right=0cm of o2_2] (o2_3) {1};
        \node[bit, right=0cm of o2_3] (o2_4) {1};
        \node[bit, right=0cm of o2_4] (o2_5) {0};
        \node[targetbit, right=0cm of o2_5] (o2_6) {0}; 
        \node[bit, right=0cm of o2_6] (o2_7) {1};
        
        \draw[arrow, color=purple!80!black] (i2_7.south) to[out=-90, in=90] node[pos=0.40, lbl, text=purple!80!black] {C1} (o2_6.north);
        \draw[arrow, color=purple!80!black] (i2_0.south) to[out=-90, in=90] node[pos=0.50, lbl, text=purple!80!black] {C2} (o2_6.north);
        \draw[arrow, color=purple!80!black] (i2_1.south) to[out=-90, in=90] node[pos=0.60, lbl, text=purple!80!black] {C3} (o2_6.north);
        \draw[arrow, color=purple!80!black] (i2_2.south) to[out=-90, in=90] node[pos=0.70, lbl, text=purple!80!black] {C4} (o2_6.north);
        \draw[arrow, color=purple!80!black] (i2_3.south) to[out=-90, in=90] node[pos=0.80, lbl, text=purple!80!black] {C5} (o2_6.north);
        \draw[arrow, color=purple!80!black] (i2_4.south) to[out=-90, in=90] node[pos=0.90, lbl, text=purple!80!black] {C6} (o2_6.north);
        \draw[arrow, color=purple!80!black] (i2_5.south) to[out=-90, in=90] node[pos=0.96, lbl, text=purple!80!black] {C7} (o2_6.north);
    \end{scope}

    \begin{scope}[xshift=9.6cm, yshift=0cm]
        \node[font=\bfseries\scriptsize, color=teal!80!black, anchor=west] at (-0.2, 3.8) {3. Left Shifts (L1-L7)};
        \node[bit] (i3_0) at (0, 3.0) {1};
        \node[bit, right=0cm of i3_0] (i3_1) {0};
        \node[bit, right=0cm of i3_1] (i3_2) {1};
        \node[bit, right=0cm of i3_2] (i3_3) {0};
        \node[bit, right=0cm of i3_3] (i3_4) {0};
        \node[bit, right=0cm of i3_4] (i3_5) {0};
        \node[bit, right=0cm of i3_5] (i3_6) {1};
        \node[bit, right=0cm of i3_6] (i3_7) {1};
        
        \node[phantomL, right=0.15cm of i3_7] (i3_P1) {0};
        \node[phantomL, right=0cm of i3_P1] (i3_P2) {0};
        \node[phantomL, right=0cm of i3_P2] (i3_P3) {0};
        \node[phantomL, right=0cm of i3_P3] (i3_P4) {0};
        \node[phantomL, right=0cm of i3_P4] (i3_P5) {0};
        \node[phantomL, right=0cm of i3_P5] (i3_P6) {0};
        
        \node[bit] (o3_0) at (0, 0.0) {1};
        \node[bit, right=0cm of o3_0] (o3_1) {1};
        \node[bit, right=0cm of o3_1] (o3_2) {0};
        \node[bit, right=0cm of o3_2] (o3_3) {1};
        \node[bit, right=0cm of o3_3] (o3_4) {1};
        \node[bit, right=0cm of o3_4] (o3_5) {0};
        \node[targetbit, right=0cm of o3_5] (o3_6) {0}; 
        \node[bit, right=0cm of o3_6] (o3_7) {1};
        
        \draw[arrow, color=teal!80!black] (i3_7.south) to[out=-90, in=90] node[pos=0.35, lbl, text=teal!80!black] {L1} (o3_6.north);
        \draw[arrow, color=teal!80!black] (i3_P1.south) to[out=-90, in=90] node[pos=0.45, lbl, text=teal!80!black] {L2} (o3_6.north);
        \draw[arrow, color=teal!80!black] (i3_P2.south) to[out=-90, in=90] node[pos=0.55, lbl, text=teal!80!black] {L3} (o3_6.north);
        \draw[arrow, color=teal!80!black] (i3_P3.south) to[out=-90, in=90] node[pos=0.65, lbl, text=teal!80!black] {L4} (o3_6.north);
        \draw[arrow, color=teal!80!black] (i3_P4.south) to[out=-90, in=90] node[pos=0.75, lbl, text=teal!80!black] {L5} (o3_6.north);
        \draw[arrow, color=teal!80!black] (i3_P5.south) to[out=-90, in=90] node[pos=0.85, lbl, text=teal!80!black] {L6} (o3_6.north);
        \draw[arrow, color=teal!80!black] (i3_P6.south) to[out=-90, in=90] node[pos=0.92, lbl, text=teal!80!black] {L7} (o3_6.north);
    \end{scope}

\end{tikzpicture}
\vspace{-0.2cm}
\caption{Spatial mapping of input bits to form the bases for target output Bit 6 (red) across the three base classes: Right Shifts (1), Circular Shifts (2), and Left Shifts (3). The curved arrows trace how each individual offset (e.g., $R_1$, $C_2$, or $L_3$) retrieves its value from a specific position in the input string or from padded zeros (dashed).}
\label{fig:receptive_fields}
\end{figure}
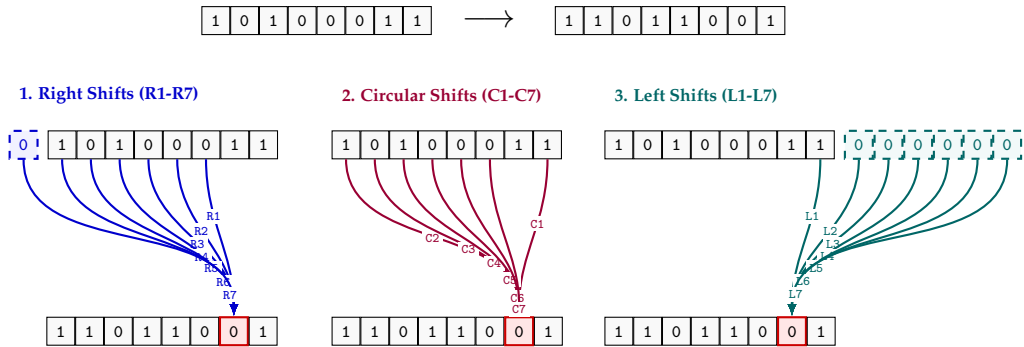

Our approach begins by deconstructing the sequence-to-sequence translation paradigm. Instead of mapping an entire 8-bit input directly to an 8-bit output, we break down a single 8-bit example into eight independent 1-bit transformations.

Predicting a single output bit $y$ at position $i$ based solely on the input bit at the exact same position $i$ is insufficient if the logical rule involves bits moving across the sequence (via shifts or rotations). To natively bypass this, we define a set of 22 spatial pointers, which we call \textbf{Bases}. These bases allow us to directly query the status of any input bit that could potentially influence our target output bit:
\begin{itemize}[leftmargin=2em, itemsep=0.1em, topsep=0.2em]
    \item \textbf{\texttt{x}} (Original): Looks directly at the input bit at the target position $i$.
    \item \textbf{\texttt{R1-R7}} (Right Shifts): Act as left-pointing spatial pointers. For instance, if the puzzle requires a Right Shift of $k$ ($R_k$), the output value at target index $i$ retrieves the input bit located $k$ positions to its left (at index $i-k$ under left-to-right indexing).
    \item \textbf{\texttt{L1-L7}} (Left Shifts): Act as right-pointing spatial pointers. For instance, a Left Shift of $k$ ($L_k$) at target index $i$ retrieves the input bit located $k$ positions to its right (at index $i+k$).
    \item \textbf{\texttt{C1-C7}} (Circular Rotations): Act as wrapping pointers. For instance, $C_k$ at target index $i$ retrieves the input bit $k$ positions to its right, wrapping around to the left side of the array if the boundary is exceeded.
\end{itemize}

As illustrated in Figure \ref{fig:receptive_fields}, these bases map any dynamic spatial movement of bits into a static set of 22 Boolean features $\{0, 1\}$ for each output position. The entire bit manipulation puzzle is thus mathematically reduced to discovering a simple boolean function $f$ mapping a subset of these bases to the target bit $y$:
\begin{equation}
f(x, R_1, \dots, R_7, C_1, \dots, C_7, L_1, \dots, L_7) = y
\end{equation}

\vspace{-0.4cm}
\begin{tcolorbox}[colback=darkblue!4,colframe=darkblue!80,fonttitle=\bfseries\small,title=\textbf{Box 1: Deconstructing an 8-bit String into Boolean Bases},arc=2pt,boxrule=0.7pt,left=5pt,right=5pt,top=4pt,bottom=4pt]
\small
Suppose we are given the transformation from Figure \ref{fig:receptive_fields}: \texttt{10100011} $\rightarrow$ \texttt{11011001}. We assign string indices from left to right (\texttt{0} to \texttt{7}). Let us isolate \textbf{Bit 6 (6th bit from left)}, which has a target output of \textbf{\texttt{0}}. To understand \textit{why}, we evaluate our 22 bases acting as spatial pointers on the input:
\begin{itemize}[leftmargin=1.5em, itemsep=0pt, topsep=2pt, parsep=0pt]
    \item \textbf{\texttt{x} (Original):} Looks at Bit 6 $\rightarrow$ \texttt{1}
    \item \textbf{\texttt{R1} (Right Shift 1):} Looks at adjacent left Bit 5 $\rightarrow$ \texttt{0}
    \item \textbf{\texttt{C1} (Circular Left 1):} Looks at adjacent right Bit 7 $\rightarrow$ \texttt{1}
    \item \textbf{\texttt{L1} (Left Shift 1):} Looks at adjacent right Bit 7 $\rightarrow$ \texttt{1}
    \item \textbf{\texttt{L3} (Left Shift 3):} Exceeds boundaries, brings in a constant $\rightarrow$ \texttt{0}
\end{itemize}
By extracting the 22 Boolean bases in order ($x$, $R_1 \dots R_7$, $C_1 \dots C_7$, $L_1 \dots L_7$) for this specific bit, we evaluate our function:
$$ \text{Bit 6: } \quad f(1, \, 0,0,0,1,0,1,0, \, 1,1,0,1,0,0,0, \, 1,0,0,0,0,0,0) = \mathbf{0} $$

Repeating this physical mapping transforms this single 8-bit string into 8 independent dataset rows:
\begin{center}
\begin{tabular}{ll}
\texttt{Bit 0:} $f(1, \dots \text{22 bases} \dots) = \mathbf{1}$ \qquad & \texttt{Bit 3:} $f(0, \dots \text{22 bases} \dots) = \mathbf{1}$ \\
\texttt{Bit 1:} $f(0, \dots \text{22 bases} \dots) = \mathbf{1}$ \qquad & \texttt{Bit 2:} $f(0, \dots \text{22 bases} \dots) = \mathbf{0}$ \\
\texttt{Bit 5:} $f(1, \dots \text{22 bases} \dots) = \mathbf{0}$ \qquad & \texttt{Bit 1:} $f(1, \dots \text{22 bases} \dots) = \mathbf{0}$ \\
\texttt{Bit 4:} $f(0, \dots \text{22 bases} \dots) = \mathbf{1}$ \qquad & \texttt{Bit 0:} $f(1, \dots \text{22 bases} \dots) = \mathbf{1}$
\end{tabular}
\end{center}
Expanded across 8 puzzle examples, the dataset becomes a structured 64-row boolean logic table.
\end{tcolorbox}

\subsection{Empirical Truth Tables}

Because our generated bases contain only binary values (\texttt{0} and \texttt{1}), any underlying logical rule—no matter how complex the combination of \texttt{AND}, \texttt{OR}, or \texttt{XOR} gates—can be perfectly represented by a simple Truth Table. This mathematical property completely eliminates the need to deduce an explicit algebraic equation.

Once we identify the specific subset of bases that dictate the output, we do not need to figure out how they are mathematically connected. We simply observe the existing dataset and record the target output for each state. This insight dramatically simplifies the problem: the task is reduced entirely to discovering \textit{which} subset of bases to look at, replacing algebraic derivation with straightforward data observation.

\begin{tcolorbox}[colback=darkblue!4,colframe=darkblue!80,fonttitle=\bfseries\small,title=\textbf{Box 2: Building an Empirical Truth Table},arc=2pt,boxrule=0.7pt,left=5pt,right=5pt,top=4pt,bottom=4pt]
\small
Suppose we know the target output is entirely controlled by two bases: the original bit (\textbf{\texttt{x}}) and a left-shifted bit (\textbf{\texttt{L1}}). Instead of trying to guess the boolean algebra formula connecting them, we simply observe the 64 rows in our dataset to see what happens in each state:

\vspace{0.1cm}
\begin{center}
\renewcommand{\arraystretch}{1.2}
\begin{tabular}{ccc|c}
\toprule
\textbf{Observed Dataset Row} & \textbf{Base \texttt{x}} & \textbf{Base \texttt{L1}} & \textbf{Target Output} \\
\midrule
Row 5 & \texttt{0} & \texttt{0} & \textbf{\texttt{0}} \\
Row 12 & \texttt{0} & \texttt{1} & \textbf{\texttt{1}} \\
Row 27 & \texttt{1} & \texttt{0} & \textbf{\texttt{1}} \\
Row 42 & \texttt{1} & \texttt{1} & \textbf{\texttt{0}} \\
\bottomrule
\end{tabular}
\end{center}
\vspace{0.1cm}

By matching the inputs to the outputs, we have successfully constructed the governing logic rule. We never needed to mathematically deduce that this is an \texttt{XOR} gate; the empirical truth table naturally provides the exact state-to-state mapping required to solve the puzzle.
\end{tcolorbox}

\subsection{Isolating Output-Influencing Bases via Minimal Bitflips}

\begin{figure}[h]
\centering
\begin{tikzpicture}[
    scale=0.88, transform shape,
    bit/.style={draw, minimum size=0.45cm, font=\ttfamily\scriptsize, fill=gray!5, inner sep=0pt},
    diffbit/.style={draw=red!80!black, thick, minimum size=0.45cm, font=\ttfamily\scriptsize\bfseries, fill=red!10, text=red!80!black, inner sep=0pt},
    outbit/.style={draw=darkblue, thick, minimum size=0.45cm, font=\ttfamily\small\bfseries, fill=darkblue!10, text=darkblue, inner sep=0pt},
    arrow/.style={-{Latex[scale=0.8]}, thick, color=red!80!black},
    lbl/.style={font=\scriptsize\bfseries, color=red!80!black}
]

    \node[font=\bfseries\small, color=darkblue, anchor=west] at (-2.0, 1.4) {Scenario A: Single Bit Flip (Absolute Certainty)};

    \matrix [matrix of nodes, nodes={bit}, column sep=-\pgflinewidth, row sep=0.15cm] (matA) at (3.5, 0.1)
    {
      1 & 0 & 1 & 0 & 0 & 1 & 1 & 1 & 1 & 0 & 1 & 0 & 1 & 0 & 0 & |[diffbit]| 1 & 1 & 1 & 1 & 0 & 0 & 0 \\
      1 & 0 & 1 & 0 & 0 & 1 & 1 & 1 & 1 & 0 & 1 & 0 & 1 & 0 & 0 & |[diffbit]| 0 & 1 & 1 & 1 & 0 & 0 & 0 \\
    };
    
    \node[left=0.2cm of matA-1-1, font=\scriptsize\color{gray}\bfseries] {Row 1:};
    \node[left=0.2cm of matA-2-1, font=\scriptsize\color{gray}\bfseries] {Row 2:};

    \node[right=0.2cm of matA-1-22, font=\bfseries] (arA1) {$\longrightarrow$};
    \node[outbit, right=0.2cm of arA1] {0};
    \node[right=0.2cm of matA-2-22, font=\bfseries] (arA2) {$\longrightarrow$};
    \node[outbit, right=0.2cm of arA2] {1};
    
    \node[above=0.1cm of matA-1-16, font=\scriptsize\bfseries\color{red!80!black}] {L1};
    
    \node[lbl, align=center, fill=white, draw=red!80!black, thick, rounded corners=3pt, inner sep=6pt] (textA) at (3.5, -1.4) {
        Only \textbf{L1} flipped $\longrightarrow$ \textbf{L1} is the Output-Influencing Base
    };
    \draw[arrow] (matA-2-16.south) to[out=-90, in=90] (textA.north);

    \node[font=\bfseries\small, color=darkblue, anchor=west] at (-2.0, -2.9) {Scenario B: Multi-Bit Flip (Under-determined Constraint)};

    \matrix [matrix of nodes, nodes={bit}, column sep=-\pgflinewidth, row sep=0.15cm] (matB) at (3.5, -4.1)
    {
      1 & 0 & 0 & 0 & 0 & 1 & 1 & 1 & |[diffbit]| 1 & 0 & 1 & 0 & 1 & 0 & 0 & |[diffbit]| 1 & 0 & 1 & 1 & 0 & 0 & 0 \\
      1 & 0 & 0 & 0 & 0 & 1 & 1 & 1 & |[diffbit]| 0 & 0 & 1 & 0 & 1 & 0 & 0 & |[diffbit]| 0 & 0 & 1 & 1 & 0 & 0 & 0 \\
    };
    
    \node[left=0.2cm of matB-1-1, font=\scriptsize\color{gray}\bfseries] {Row 3:};
    \node[left=0.2cm of matB-2-1, font=\scriptsize\color{gray}\bfseries] {Row 4:};

    \node[right=0.2cm of matB-1-22, font=\bfseries] (arB1) {$\longrightarrow$};
    \node[outbit, right=0.2cm of arB1] {0};
    \node[right=0.2cm of matB-2-22, font=\bfseries] (arB2) {$\longrightarrow$};
    \node[outbit, right=0.2cm of arB2] {1};
    
    \node[above=0.1cm of matB-1-9, font=\scriptsize\bfseries\color{red!80!black}] {C1};
    \node[above=0.1cm of matB-1-16, font=\scriptsize\bfseries\color{red!80!black}] {L1};
    
    \node[lbl, align=center, fill=white, draw=red!80!black, thick, rounded corners=3pt, inner sep=6pt] (textB) at (3.5, -5.8) {
        \textbf{C1 and L1 flipped. Three Possibilities:}\\
        1) \textbf{C1} solely caused output change \quad 2) \textbf{L1} solely caused output change \quad 3) \textbf{Both} are influencing bases
    };
    
    \draw[arrow] (matB-2-9.south) to[out=-90, in=90] ($ (textB.north) + (-2.2cm, 0) $);
    \draw[arrow] (matB-2-16.south) to[out=-90, in=90] ($ (textB.north) + (2.2cm, 0) $);

\end{tikzpicture}
\vspace{-0.2cm}
\caption{Isolating output-influencing bases by comparing rows with different target outputs. When two nearly identical 22-base arrays yield different outputs (\texttt{0} vs. \texttt{1}), the differing bases must be responsible for the state change.}
\label{fig:minimal_bitflips}
\end{figure}
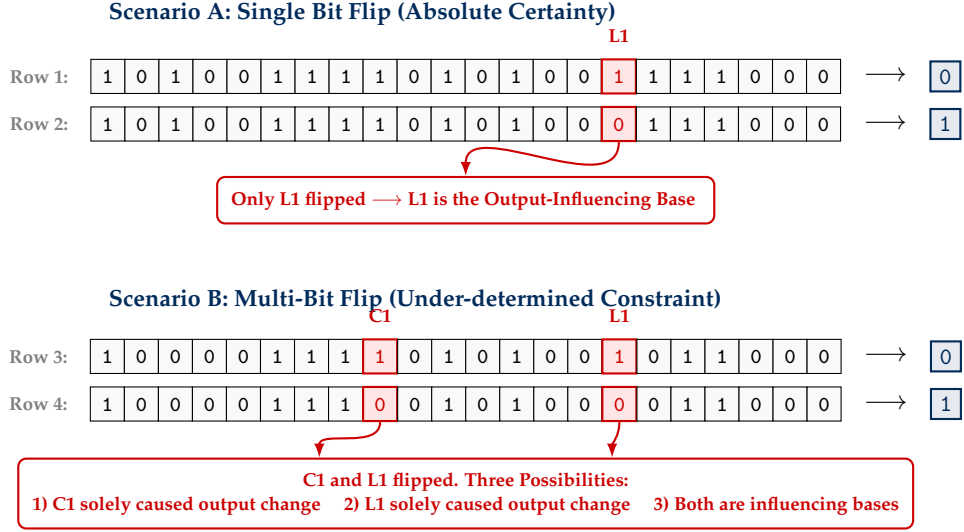

In our expanded 22-dimensional Boolean base space, asking an Auto-Regressive LLM to natively deduce the correct logical mapping is highly prone to hallucination. We bypass this spatial arithmetic calculation entirely by leveraging a foundational principle of logic and string similarity: \textit{If two nearly identical base configurations yield different outputs, the cause of the output change must lie strictly within the bases that differ.}

By splitting the 64 target output bits into two classes—outputs resulting in \texttt{0} and \texttt{1}—we systematically compare their 22-base arrays. We map every \texttt{0}-output row to a \texttt{1}-output row with the smallest Hamming distance to isolate the ``Minimal BitFlips.'' These differing bits act as strict logical constraints, which we refer to as \textbf{Flip Traces}.

As illustrated in Figure \ref{fig:minimal_bitflips}, this string-matching comparison yields two types of logical constraints:
\begin{itemize}[leftmargin=1.5em, itemsep=0.2em, topsep=0.2em]
    \item \textbf{Single-Bit Flips (Mandatory Bases):} If two rows differ in their output, but only \textit{one} base is different between them (a Hamming distance of 1), we have absolute mathematical certainty. That flipped base must be the reason the output changed. It is locked in as a mandatory base.
    \item \textbf{Multi-Bit Flips (Logical Constraints):} If two rows differ in their output and \textit{multiple} bases differ, logic dictates that the change in output was caused by one of those flipped bases, or a combination of them. This trace acts as an explicit logical OR constraint.
\end{itemize}

To ensure a discovered logical rule is universally true, it must satisfy every single Flip Trace constraint generated from the dataset. This framework perfectly translates the bit manipulation task into the classic \textbf{Set Cover} problem: \textit{What is the absolute smallest combination of bases required to ``cover'' at least one base in every trace?} By applying a Set Cover search across the entire global dataset, we can resolve these under-determined constraints and guarantee the selection of the minimal, optimal combination of bases—filtering out spurious variables without executing a single arithmetic calculation.


\section{Solver Algorithm}

We now translate our theoretical framework into an algorithm that completely solves the puzzle from start to finish. The overarching goal of this pipeline is to unify string matching and structured search to autonomously deduce the hidden logic rule.

\begin{figure}[h]
\centering
\begin{tikzpicture}[
    node distance=0.8cm and 1.8cm, 
    scale=0.78, transform shape,  
    box/.style={draw=darkblue, thick, fill=blue!5, align=center, rounded corners=2pt, font=\scriptsize\bfseries, minimum width=3.5cm, inner sep=6pt},
    loopbox/.style={draw=red!80!black, thick, fill=red!5, align=center, rounded corners=2pt, font=\scriptsize\bfseries, minimum width=3.5cm, inner sep=6pt},
    outbox/.style={draw=green!60!black, thick, fill=green!5, align=center, rounded corners=2pt, font=\scriptsize\bfseries, minimum width=3.5cm, inner sep=6pt},
    arrow/.style={-{Latex[scale=0.8]}, thick, color=darkblue},
    backarrow/.style={-{Latex[scale=0.8]}, thick, dashed, color=red!80!black},
    lbl/.style={font=\tiny\bfseries, fill=white, inner sep=1.5pt, rounded corners=2pt, align=center}
]

    \node[box] (data) {1. Row Deconstruction\\(64 Rows $\times$ 22 Bases)};
    \node[box, below=0.7cm of data] (match) {2. String Matching\\(Compare Output \texttt{0} vs. \texttt{1})};
    \node[box, below=0.7cm of match] (traces) {3. Constraint Formulation\\(Extract Flip Traces)};
    
    \node[loopbox, right=2.0cm of match] (dfs) {4. DFS Backtracking\\(Propose Base Subset)};
    \node[loopbox] (oracle) at (dfs |- traces) {5. Global Collision Check\\(Test across all 64 rows)};
    
    \node[outbox, right=2.0cm of dfs] (tt) {6. Synthesize Truth Table};
    \node[outbox] (predict) at (tt |- oracle) {7. Target Prediction\\(Solve Unseen Input)};

    \draw[arrow] (data.south) -- (match.north);
    \draw[arrow] (match.south) -- (traces.north);
    
    \draw[arrow] (traces.east) -- ++(0.3,0) |- (dfs.west);
    \node[lbl, text=darkblue, align=center, anchor=south] at ($(dfs.west) + (-0.8, 0.05)$) {Universe of\\Constraints};
    
    \draw[arrow] (dfs.south) -- node[right, lbl, text=darkblue] {Candidate} (oracle.north);
    \draw[backarrow] (oracle.west) to[bend left=32] (dfs.west);
    \node[lbl, text=red!80!black, align=center, anchor=east, xshift=-0.05cm] at ($(dfs.west)!0.5!(oracle.west) + (-0.35, 0)$) {Collision!\\(Backtrack)};
    
    \draw[arrow, color=green!50!black] (oracle.east) -- ++(0.3,0) |- (tt.west);
    \node[lbl, text=green!50!black, align=center, anchor=south] at ($(tt.west) + (-0.8, 0.05)$) {No Collisions\\(Valid)};
    
    \draw[arrow, color=green!50!black] (tt.south) -- (predict.north);

    \begin{pgfonlayer}{background}
        \filldraw[fill=gray!10, draw=gray!40, dashed, rounded corners] 
            ($(data.north west)+(-0.3,0.3)$) rectangle ($(traces.south east)+(0.3,-0.3)$);
        \node[font=\scriptsize\bfseries\color{gray}, anchor=north] at ($(data.north)+(0,0.6)$) {Phase 1: String Matching};

        \filldraw[fill=red!2, draw=red!30, dashed, rounded corners] 
            ($(dfs.north west)+(-1.1,0.3)$) rectangle ($(oracle.south east)+(0.3,-0.3)$);
        \node[font=\scriptsize\bfseries\color{red!60!black}, anchor=north] at ($(dfs.north)+(-0.4,0.6)$) {Phase 2: Search \& Verification};
        
        \filldraw[fill=green!2, draw=green!30, dashed, rounded corners] 
            ($(tt.north west)+(-0.3,0.3)$) rectangle ($(predict.south east)+(0.3,-0.3)$);
        \node[font=\scriptsize\bfseries\color{green!60!black}, anchor=north] at ($(tt.north)+(0,0.6)$) {Phase 3: Synthesis};
    \end{pgfonlayer}

\end{tikzpicture}
\vspace{-0.2cm}
\caption{End-to-End Flowchart of the Bit Manipulation Solver. The algorithm isolates logical constraints via string matching, iteratively searches for valid bases via backtracking, and ultimately synthesizes the final truth table to solve the puzzle.}
\label{fig:flowchart}
\end{figure}
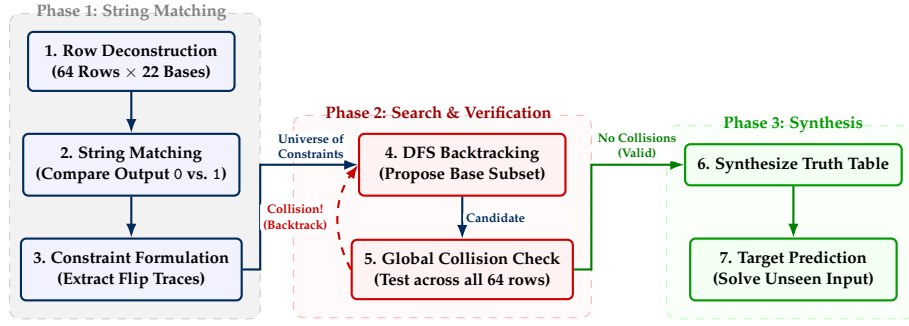

As visualized in Figure \ref{fig:flowchart}, the solver operates in three distinct phases: first, it uses string similarity to find the logical constraints; second, it employs a ``guess-and-check'' backtracking search to find a perfectly valid subset of bases; and finally, it synthesizes the empirical truth table to predict the target string. 

The specific execution steps of this pipeline are formalized in \textbf{Algorithm 1}. 

\begin{tcolorbox}[colback=darkblue!2,colframe=darkblue!80,fonttitle=\bfseries\small,title=\textbf{Algorithm 1: End-to-End Bit Manipulation Solver},arc=2pt,boxrule=0.7pt,left=5pt,right=5pt,top=4pt,bottom=4pt]
\small
\textbf{Input:} Puzzle Examples, Unseen Target String \\
\textbf{Output:} 8-bit Target Prediction
\vspace{0.3em}\hrule\vspace{0.3em}
\begin{tabbing}
\textbf{Phase 1: Feature Extraction \& String Matching} \\
1. \= \textbf{Create Bases:} Generate the 22 spatial bases ($\mathcal{B}$) for all 8-bit inputs. \\
2. \> \textbf{Row Deconstruction:} Flatten the inputs into a 64-row dataset. \\
3. \> \textbf{String Matching:} Compare rows outputting \texttt{0} against rows outputting \texttt{1}.\\
   \> \quad $\rightarrow$ Extract the differing bases to create our \textbf{Flip Traces}. \\[0.5em]

\textbf{Phase 2: Backtracking DFS \& Collision Verification} \\
4. \> \textbf{Lock Mandatory Bases:} If a trace has only 1 flipped base, permanently lock it. \\
5. \> \textbf{Function} \texttt{DFS\_Search}(\textit{Current\_Bases}, \textit{Uncovered\_Traces}): \\
   \> \quad \= \textbf{If} all traces are covered: \\
   \> \quad \quad \= Evaluate \textit{Current\_Bases} across all 64 rows to build a Truth Table. \\
   \> \quad \quad \> \textbf{If} two identical input states yield different outputs $\rightarrow$ \textbf{Collision Detected!} \\
   \> \quad \quad \> \quad \textbf{Return} False \quad \textit{(Reject candidate, trigger backtrack)} \\
   \> \quad \quad \> \textbf{Else}: \textbf{Return} \textit{Current\_Bases} \quad \textit{(Mathematically valid bases found!)} \\
   \> \quad \textbf{Rank} remaining bases in \textit{Uncovered\_Traces} by occurrence frequency. \\
   \> \quad \textbf{For} each candidate base $b$: \\
   \> \quad \quad \> Add $b$ to \textit{Current\_Bases}, update \textit{Uncovered\_Traces}. \\
   \> \quad \quad \> \textbf{If} \texttt{DFS\_Search}(\textit{Current\_Bases}, \textit{Uncovered\_Traces}) is valid: \textbf{Return} result.\\
   \> \quad \textbf{Return} False \quad \textit{(Exhausted branch, backtrack to previous state)}\\[0.5em]

\textbf{Phase 3: Rule Synthesis \& Target Prediction} \\
6. \> \textbf{Synthesize:} Use the verified \textit{Current\_Bases} to finalize the exact Truth Table. \\
7. \> \textbf{Solve:} Apply the bases and Truth Table to the unseen Target Input String.
\end{tabbing}
\end{tcolorbox}

\subsection{Phase 1: Feature Extraction and String Matching}
The algorithm begins by generating the 22 bases for all 64 independent bits in the puzzle examples. To figure out which bases actually influence the output, we leverage simple string similarity. 

\begin{tcolorbox}[colback=darkblue!4,colframe=darkblue!80,fonttitle=\bfseries\small,title=\textbf{Box 2: Formulating the Constraint Universe},arc=2pt,boxrule=0.7pt,left=5pt,right=5pt,top=4pt,bottom=4pt]
\small
Suppose after comparing the 64 rows of a puzzle, our string-matching phase extracts three unique Flip Traces:
\begin{itemize}[leftmargin=1.5em, itemsep=4pt, topsep=2pt, parsep=0pt]
    \item \textbf{Trace 1:} \texttt{[L1]} $\longrightarrow$ \textit{An output bit changed from 0 to 1 while only the L1 base flipped, making L1 a mandatory base with absolute certainty.}
    \item \textbf{Trace 2:} \texttt{[L1, C1]} $\longrightarrow$ \textit{Another output bit changed from 0 to 1 while both L1 and C1 flipped; thus, either L1, C1, or both caused the change.}
    \item \textbf{Trace 3:} \texttt{[C1, R4]} $\longrightarrow$ \textit{A third output bit changed from 0 to 1 while C1 and R4 flipped, establishing a logical OR constraint where at least one must be selected.}
\end{itemize}
\vspace{0.1cm}
Before searching, the algorithm counts the occurrence frequency of each base across these traces. In this toy universe, \textbf{\texttt{L1}} appears twice, \textbf{\texttt{C1}} appears twice, and \textbf{\texttt{R4}} appears once.
\end{tcolorbox}

By comparing the rows that output \texttt{0} with the rows that output \texttt{1}, the algorithm identifies the minimal structural differences required to trigger a state change. These differing bases are extracted as \textbf{Flip Traces} (our ``clues''). These traces serve as the absolute logical constraints of the puzzle: any valid Boolean rule \textit{must} select at least one base from every trace to mathematically explain why the outputs changed. This naturally frames the next step as a classic Set Cover problem, as demonstrated in Box 2.

\subsection{Phase 2: Backtracking Search and Collision Verification}
With our list of constraints defined, the algorithm must find the absolute smallest combination of bases that ``covers'' every trace. We execute a structured search process.

The search utilizes \textbf{frequency-guided depth first search}. When the algorithm encounters traces with multiple possibilities, it ranks the bases by how often they appear globally. It greedily ``guesses'' the most frequent bases first, naturally favoring the simplest logical rules. 

However, simply finding a set of bases that covers our traces is not enough; we must verify against all the 64 rows. To prove the rule is universally sound, the algorithm performs a \textbf{Global Collision Verification}. It evaluates the currently guessed bases across \textit{all 64 rows} of the global dataset. 

\begin{figure}[h]
\centering
\begin{tikzpicture}[
    scale=0.85, transform shape,
    treenode/.style = {align=center, inner sep=6pt, font=\ttfamily\scriptsize, draw=darkblue, thick, fill=blue!5, rounded corners=3pt},
    colnode/.style = {align=center, inner sep=6pt, font=\ttfamily\scriptsize, draw=red!80!black, thick, fill=red!5, rounded corners=3pt},
    succnode/.style = {align=center, inner sep=6pt, font=\ttfamily\scriptsize, draw=green!60!black, thick, fill=green!5, rounded corners=3pt},
    oraclenode/.style = {align=center, inner sep=4pt, font=\scriptsize, draw=gray, thick, fill=white, rounded corners=2pt},
    arrow/.style = {thick, -{Latex[scale=0.8]}, color=darkblue},
    backarrow/.style = {thick, dashed, -{Latex[scale=0.8]}, color=red!80!black},
    lbl/.style = {font=\scriptsize\bfseries, color=darkblue, fill=white, inner sep=2pt}
]

    \node [treenode] (root) {\textbf{Start DFS Search}\\Locked Base: \textbf{\texttt{L1}} \textit{(Covers Traces 1 \& 2)}\\Uncovered Traces: \textbf{\texttt{T3 [C1, R4]}}};

    \node [colnode, below=1.0cm of root, xshift=-3.6cm] (C1) {\textbf{Guess 1:} Add \textbf{\texttt{C1}} (Freq: 2)\\Candidate Set: \textbf{\texttt{\{L1, C1\}}}};
    \node [succnode, below=1.0cm of root, xshift=3.6cm] (R4) {\textbf{Guess 2:} Add \textbf{\texttt{R4}} (Freq: 1)\\Candidate Set: \textbf{\texttt{\{L1, R4\}}}};

    \node [oraclenode, draw=red!80!black, below=0.4cm of C1] (evalC1) {
        \textbf{Global Oracle Check} \\[4pt]
        \renewcommand{\arraystretch}{1.1}
        \begin{tabular}{cc|c|l}
        \texttt{L1} & \texttt{C1} & \textbf{Out} & \textbf{Row} \\ \hline
        1 & 0 & \texttt{0} & 2 \\
        1 & 0 & \textcolor{red}{\textbf{\texttt{1}}} & \textcolor{red}{\textbf{15}} \\
        \end{tabular} \\[4pt]
        \textcolor{red!80!black}{\textbf{Collision Detected! (Reject)}}
    };

    \node [oraclenode, draw=green!60!black, below=0.4cm of R4] (evalR4) {
        \textbf{Global Oracle Check} \\[4pt]
        \renewcommand{\arraystretch}{1.1}
        \begin{tabular}{cc|c|l}
        \texttt{L1} & \texttt{R4} & \textbf{Out} & \textbf{Row} \\ \hline
        1 & 0 & \texttt{1} & 5 \\
        1 & 0 & \texttt{1} & 12 \\
        \multicolumn{4}{c}{\textit{... (Consistent across 64 rows)}}
        \end{tabular} \\[4pt]
        \textcolor{green!50!black}{\textbf{Valid Truth Table Synthesized!}}
    };

    \draw [arrow] (root.south) -- node[above left, lbl, xshift=0.1cm] {Branch 1} (C1.north);
    \draw [arrow] (root.south) -- node[above right, lbl, xshift=-0.1cm] {Branch 2} (R4.north);
    
    \draw [arrow, color=red!80!black] (C1.south) -- (evalC1.north);
    \draw [arrow, color=green!50!black] (R4.south) -- (evalR4.north);

    \draw [backarrow] (evalC1.west) to[out=180, in=180, looseness=1.2] node[left, font=\scriptsize\bfseries, text=red!80!black] {Backtrack} (root.west);

\end{tikzpicture}
\vspace{-0.2cm}
\caption{Visualizing the Phase 2 Backtracking Search using the traces from Box 2. The algorithm proposes a candidate subset of bases and tests it globally. On Branch 1, an empirical truth table reveals a logical contradiction (inputs \texttt{1,0} yielding conflicting outputs), forcing the algorithm to deterministically backtrack to Branch 2.}
\label{fig:dfs_tree}
\end{figure}
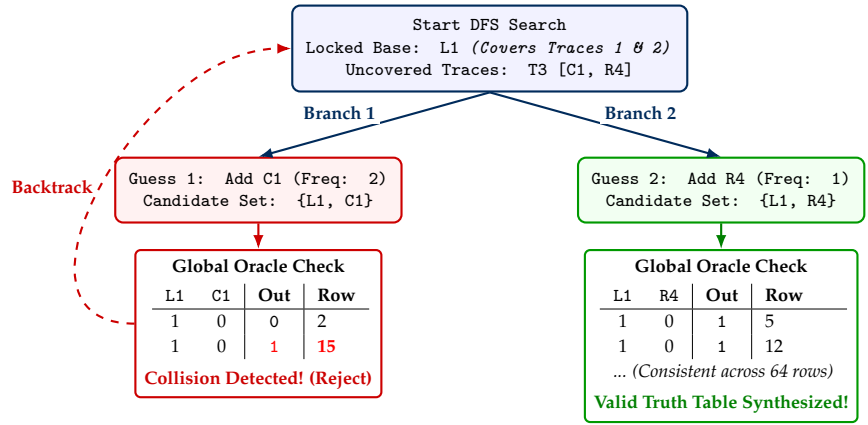

If two different rows in the dataset possess the exact same Boolean values for our chosen bases but demand conflicting target outputs (e.g., Row 2 says the output should be \texttt{0}, but Row 15 says the output should be \texttt{1}), a \textbf{Collision} has occurred. Because a function cannot map identical inputs to different outputs, a collision acts as an objective mathematical oracle. It proves our current guess is incomplete. Upon detecting a collision, the algorithm immediately rejects the subset and \textbf{backtracks} up the search tree to try a different combination of bases, ensuring strict error recovery (visualized in Figure \ref{fig:dfs_tree}).

\begin{tcolorbox}[colback=darkblue!4,colframe=darkblue!80,fonttitle=\bfseries\small,title=\textbf{Box 3: The Backtracking Execution},arc=2pt,boxrule=0.7pt,left=5pt,right=5pt,top=4pt,bottom=4pt]
\small
Following the traces from Box 2, here is how the algorithm navigates the search tree (Figure \ref{fig:dfs_tree}, Hypothetical Execution):
\begin{enumerate}[leftmargin=1.5em, itemsep=0pt, topsep=2pt, parsep=0pt]
    \item \textbf{Mandatory Lock:} Trace 1 \texttt{[L1]} is an absolute certainty. The algorithm locks \textbf{\texttt{L1}} into the final rule. Because \texttt{L1} is chosen, Trace 2 \texttt{[L1, C1]} is automatically satisfied and covered.
    \item \textbf{Greedy Guess:} The only uncovered trace is Trace 3 \texttt{[C1, R4]}. Because \texttt{C1} appeared more frequently in our initial count across all traces, the algorithm guesses it first, testing the candidate combination \textbf{\texttt{[L1, C1]}}.
    \item \textbf{The Collision (Oracle):} The algorithm attempts to build a 64-row truth table for \texttt{[L1, C1]}. It discovers a paradox: on Row 2, the inputs are $L_1\!=\!1, C_1\!=\!0$ and the target output is \textbf{\texttt{0}}. However, on Row 15, the exact same inputs ($L_1\!=\!1, C_1\!=\!0$) demand an output of \textbf{\texttt{1}}. A function cannot map identical inputs to different outputs. 
    \item \textbf{Backtrack \& Success:} The algorithm rejects \texttt{C1}, backtracks, and tries the next option: \textbf{\texttt{R4}}. It tests the candidate \textbf{\texttt{[L1, R4]}} against all 64 rows, finds zero contradictions, and locks it in as the final answer.
\end{enumerate}
\end{tcolorbox}

\subsection{Phase 3: Truth Table Synthesis and Target Prediction}
Once the backtracking algorithm successfully discovers a subset of bases that covers all traces \textit{and} passes the 64-row collision check without any contradictions, the search concludes. The hypothetical mapping generated during the verification step is now mathematically proven to be the universally correct Truth Table.

To complete the puzzle, the algorithm processes the unseen target input string. It generates the 22 spatial bases for the target and isolates the specific optimal bases discovered by our search. By querying our proven Truth Table sequentially from Bit 0 up to Bit 7, it constructs the final 8-bit output prediction, flawlessly completing the puzzle.

\section{Interactive Reasoning SFT and Tokenization}

While Section 3 describes a deterministic and mathematically rigorous solver, standard LLMs cannot natively execute Python algorithms. The final challenge is embedding this search-and-verify logic into the model's parametric memory. We achieve this through two machine learning innovations: Strict Bit Tokenization and Interactive Reasoning via Dynamic Masking.

\subsection{Overcoming Spatial Bias via Strict CoT Token Generation}

In sequence-to-sequence bit manipulation, the model's attention mechanism relies heavily on the strict spatial alignment of bits. However, standard Byte-Pair Encoding (BPE) tokenizers are optimized for natural language and frequently merge adjacent numbers. For example, an 8-bit string like \texttt{10100011} might be arbitrarily chunked into tokens like \texttt{[1010]}, \texttt{[00]}, and \texttt{[11]}. This arbitrary chunking destroys the 2D spatial grid required to compare 64 dataset rows horizontally.

Normally, this is resolved by modifying the tokenizer rules. However, the competition constraints restricted submissions strictly to Low-Rank Adaptation (LoRA) weights, meaning the base tokenizer could not be modified. The input prompt would inevitably suffer from arbitrary BPE chunking.

To bypass this architectural constraint, we engineered a workaround entirely within the training data. While we could not control how the model \textit{read} the input prompt, we could control how it \textit{generated} its reasoning. During the Supervised Fine-Tuning (SFT) data packing phase, we implemented a custom script that bypassed standard tokenization for the Chain-of-Thought (CoT) and final answer. By using regular expressions to isolate binary sequences, we manually mapped every \texttt{0} and \texttt{1} to their individual, single-character token IDs. 

By training the LoRA adapter on these un-chunked sequences, the model learned a strict generative behavior: it outputs its empirical truth tables and backtracking traces exactly one bit at a time. This forces the model to construct a perfect, un-chunked spatial grid within its own generated context window, allowing its self-attention mechanism to correctly perform string matching and feature selection despite the flawed tokenization of the original input.

\subsection{Interactive Reasoning SFT via Dynamic Masking}

Developing robust error recovery—the ability to recognize a logical failure and backtrack to an alternative hypothesis—is a critical component of reasoning. Typically, teaching models this behavior requires computationally expensive Reinforcement Learning (RL) frameworks. Operating under strict compute constraints (utilizing only contest-provided hardware), we engineered a highly efficient alternative to instill this capability purely within standard offline Supervised Fine-Tuning (SFT).

\begin{tcolorbox}[colback=darkblue!4,colframe=darkblue!80,fonttitle=\bfseries\small,title=\textbf{Box 4: Simulating Environmental Feedback via Dynamic Masking},arc=2pt,boxrule=0.7pt,left=5pt,right=5pt,top=4pt,bottom=4pt]
\small
During training, we synthetically generate the entire DFS Backtracking trace. We dynamically apply a binary mask to the tokens to simulate a live, interactive environment between the model and the Oracle.

\vspace{0.1cm}
\begin{tcolorbox}[colback=green!5, colframe=green!40!black, boxrule=0.5pt, arc=1pt, left=4pt, right=4pt, top=2pt, bottom=2pt]
\textbf{\textcolor{green!40!black}{LLM Hypothesis (Unmasked - Loss Computed):}}\\
\texttt{Uncovered Traces: T3 [C1, R4].\\
Attempting most frequent base: C1.\\
Candidate Set: \{L1, C1\}. Requesting evaluation...}
\end{tcolorbox}

\vspace{-0.1cm}
\begin{tcolorbox}[colback=gray!10, colframe=gray!50, boxrule=0.5pt, arc=1pt, left=4pt, right=4pt, top=2pt, bottom=2pt]
\textbf{\textcolor{gray!80!black}{Simulated Oracle Feedback (Masked - Loss Ignored):}}\\
\texttt{Evaluating against hidden 64-row global table:\\
| L1 | C1 || Out | Row |\\
| 1  | 0  || 0   | 2   |\\
| 1  | 0  || 1   | 15  |\\
Collision detected! (\{L1, C1\} is aliased/spurious).}
\end{tcolorbox}

\vspace{-0.1cm}
\begin{tcolorbox}[colback=green!5, colframe=green!40!black, boxrule=0.5pt, arc=1pt, left=4pt, right=4pt, top=2pt, bottom=2pt]
\textbf{\textcolor{green!40!black}{LLM Recovery (Unmasked - Loss Computed):}}\\
\texttt{Collision confirmed. Rejecting C1.\\
Backtracking... (BT \#1)\\
Attempting next base: R4.}
\end{tcolorbox}
\end{tcolorbox}

\vspace{0.2cm}
Instead of forcing the LLM to internally compute and memorize the massive 64-row global dataset to check its own work (which is computationally wasteful and prone to hallucination), we simulated an interactive environment. During training data generation, whenever the LLM proposes a candidate subset of bases, an automated external ``Oracle'' evaluates the guess against the true, hidden 64-row dataset. The Oracle then explicitly injects a prompt back into the context window, informing the LLM whether the guess was flawless or if a collision occurred (Box 4).

To execute the interaction shown in Box 4 within a single offline SFT loop, we must modify the loss objective. The standard auto-regressive SFT loss applies equally to all generated tokens:
\begin{equation}
\mathcal{L}_{\text{SFT}} = - \sum_{i=1}^{N} \log P(x_i \mid x_{<i})
\end{equation}
Using this standard equation would force the model to attempt to predict the Oracle's collision table, blending the hidden environment into the model's own reasoning trace. To prevent this, we introduce a token-level binary mask, $m_i \in \{0, 1\}$, resulting in a \textbf{Dynamic Masking} loss function:
\begin{equation}
\mathcal{L}_{\text{Interactive}} = - \sum_{i=1}^{N} m_i \log P(x_i \mid x_{<i})
\end{equation}

During dataset construction, we set $m_i = 1$ for the model's active hypotheses and recovery steps (the green boxes), and $m_i = 0$ for the Oracle's deterministic feedback (the gray boxes). By masking the environmental feedback, the LLM is completely relieved of the computational burden of evaluating the 64-row dataset. It only computes loss on its ability to propose a valid subset and, crucially, its ability to ``read'' the un-penalized Oracle tokens and logically backtrack. This highly token-efficient paradigm successfully teaches the model the advanced error-recovery mechanics of RL at a fraction of the compute cost.

\section{Experimental Results and Evaluation}

To validate the efficacy of our base-selection formulation and Interactive Reasoning SFT, we evaluated both the theoretical upper bound of our deterministic solver and the final inferential performance of the fine-tuned LLM. 

It is important to note the extreme logistical constraints under which these results were achieved. Entering the competition in the final two weeks, all training and inference were conducted strictly within the 30-hour per-participant GPU compute budget provided by the Kaggle platform. We trained for a total of approximately 1,200 steps while manually checkpointing and resuming to navigate the platform's 12-hour session limits.

\subsection{Algorithmic Upper Bound and Deterministic Limits}
Prior to training the LLM, we evaluated our Python-based deterministic solver across a rigorous validation set of $1,602$ bit manipulation puzzles to establish the theoretical upper limit of our formulation. The algorithm achieved a global accuracy of \textbf{98.63\%} ($1,580 / 1,602$). 

An algorithmic autopsy of the 22 unsolved puzzles revealed that these failures were not due to any structural limitation in our 22-base formulation (as $K > 3$ failures accounted for \textbf{0\%} of errors). Instead, these puzzles represent \textbf{objectively under-determined systems where the ground-truth rule cannot be mathematically deduced from the provided examples}:
\begin{itemize}[leftmargin=1.5em, itemsep=0.2em, topsep=0.2em]
    \item \textbf{Out-Of-Distribution (OOD) Target States (27.3\%):} In 6 puzzles, the target input required evaluating a Boolean state that was completely absent from the 8 provided examples. Because the required transition logic was never demonstrated, solving the puzzle is mathematically impossible without blind guessing.
    \item \textbf{Spurious Correlation and Aliasing (72.7\%):} In 16 puzzles, the sparse 64-row example set was mathematically insufficient to isolate a unique rule. Multiple distinct, valid rules perfectly satisfied the examples but disagreed on the unseen target string. Because the solver has no way to read the dataset creator's mind, choosing the "incorrect" rule is an inevitable artifact of an under-determined constraint space.
\end{itemize}

\subsection{LLM Performance and Ablation Study}
To evaluate the model's ability to internalize this algorithm, we conducted an ablation study comparing two sequential training regimes. Our first model was fine-tuned exclusively on our synthetic DFS traces (\textbf{Synthetic Only}). For our second model, we took the last checkpoint of this synthetic-only training and continued fine-tuning it exclusively on the competition's original dataset (\textbf{Synthetic + Original}). 

As detailed in Table \ref{tab:results_comparison}, the Synthetic Only model achieved an outstanding \textbf{96.13\%} global accuracy, successfully retaining nearly the entire capability of our deterministic solver. 

\begin{table}[h]
\centering
\renewcommand{\arraystretch}{1.2}
\resizebox{0.85\textwidth}{!}{
\begin{tabular}{@{}lc|cc|cc@{}}
\toprule
 & & \multicolumn{2}{c|}{\textbf{Synthetic Only}} & \multicolumn{2}{c}{\textbf{Synthetic + Original}} \\
\textbf{Complexity} & \textbf{Count} & \textbf{Accuracy} & \textbf{Avg Tok.} & \textbf{Accuracy} & \textbf{Avg Tok.} \\ \midrule
$K=1$ Base & 154  & 94.16\% & 4868 & 94.16\% & 4864 \\
$K=2$ Bases & 898  & 97.88\% & 4879 & 98.44\% & 4901 \\
$K=3$ Bases & 550  & 93.82\% & 5347 & 88.55\% & 5567 \\ \midrule
\textbf{Overall} & \textbf{1602} & \textbf{96.13\%} & \textbf{5039} & \textbf{94.63\%} & \textbf{5126} \\ \bottomrule
\end{tabular}
}
\vspace{0.15cm}
\caption{Performance evaluation comparing models fine-tuned exclusively on Synthetically Generated Puzzles versus continued fine-tuning with Original Data. Metrics include categorical accuracy based on number of bases ($K$) and average token generation.}
\label{tab:results_comparison}
\end{table}

Interestingly, introducing the original dataset slightly degraded the Bit Manipulation accuracy to 94.63\%. This drift was a consequence of our late entry into the competition; to maximize our overall multi-category score, we further fine-tuned the final synthetic checkpoint on the original competition data in the final minutes, completing a mere 400 steps. While this final training step yielded our absolute best performance across all combined competition categories, it caused a minor task-specific drift in Bit Manipulation. Mixing in standard, unstructured reasoning data diluted the model's strict, token-efficient generative behavior, forcing it into longer backtracking loops (as reflected by the increased average token metrics). We hypothesize that given more training steps and a stabilized, longer fine-tuning schedule, the model would resolve this drift and smoothly converge toward the theoretical ceiling.

\subsection{LLM Backtracking Search Efficiency}
The primary objective of our Interactive Reasoning SFT was to instill System-2 error recovery, enabling the LLM to autonomously navigate dead-ends. To verify if the model successfully learned this behavior, we tracked the exact number of backtracking (BT) steps executed by the \textit{Synthetic Only} model during inference (Table \ref{tab:backtracking_llm}).

\begin{table}[h]
\centering
\renewcommand{\arraystretch}{1.2}
\resizebox{0.9\textwidth}{!}{
\begin{tabular}{@{}l|ccccccc|c@{}}
\toprule
\textbf{Complexity} & \textbf{0 BT} & \textbf{1 BT} & \textbf{2 BT} & \textbf{3 BT} & \textbf{4 BT} & \textbf{5 BT} & \textbf{$>$5 BT} & \textbf{Total} \\ \midrule
$K=1$ Base & 148 & 0 & 0 & 0 & 0 & 0 & 0 & \textbf{148} \\
$K=2$ Bases & 787 & 30 & 15 & 2 & 6 & 14 & 5 & \textbf{859} \\
$K=3$ Bases & 360 & 64 & 9 & 26 & 21 & 9 & 78 & \textbf{567} \\ \midrule
\textbf{Overall} & \textbf{1295} & \textbf{94} & \textbf{24} & \textbf{28} & \textbf{27} & \textbf{23} & \textbf{83} & \textbf{1574} \\
\bottomrule
\end{tabular}
}
\caption{Distribution of backtracking (BT) steps autonomously executed by the fine-tuned LLM during inference, categorized by the spatial base complexity ($K$) of the puzzle.}
\label{tab:backtracking_llm}
\end{table}

The results provide empirical proof that the dynamic masking strategy was highly effective. While straightforward 1-base rules were solved instantly (0 backtracks), complex $K=3$ rules frequently forced the model down aliased, spurious branches. Remarkably, the model successfully recovered from these logical collisions, executing 5 or more successive backtracks on 83 different puzzles. This proves that the LLM did not merely memorize successful paths during training; it actively utilized collision check to dynamically traverse the Set Cover search tree.

\subsection{LLM Error Diagnosis}
To further decouple reasoning failures from architectural limitations, we isolated the 62 failed predictions from our best-performing Synthetic Only model. 

The diagnosis revealed that \textbf{48.4\% (30 puzzles) of the model's errors were caused by Context Limit Truncation.} Because highly deceptive puzzles require extensive backtracking, the model's generative trace exceeded $\sim \! 7,500$ tokens. The model was architecturally cut off before it could print the final \texttt{\char`\\boxed\{\}} prediction. 29 puzzles (46.8\% of errors) were due to true LLM hallucination or failing to navigate the search tree properly. 

These metrics indicate that the underlying search algorithm and Interactive Reasoning SFT are exceptionally robust. Despite our strict hardware constraints—learning to solve the dataset in only 1,200 total training steps on limited Kaggle compute—our approach reached a remarkable 96.13\% accuracy. Given more training steps, we expect the approach to effortlessly reach its deterministic theoretical ceiling of 98.63\%.

\section{Conclusion}

In this paper, we demonstrated that Large Language Models can overcome severe combinatorial explosions in bit manipulation tasks when the underlying problem is mathematically reformulated. By converting bitwise arithmetic into a discrete base-selection and string-matching puzzle, and by enforcing strict single-bit tokenization, we eliminated the spatial alignment biases that traditionally trigger arithmetic hallucinations. 

Furthermore, we addressed the inherent difficulty Auto-Regressive models face when attempting to natively recognize and recover from their own logical errors. By introducing Interactive Reasoning SFT via Dynamic Masking, we taught the model the mechanics of hypothesis generation, simulated feedback evaluation, and deterministic backtracking. 

Despite entering the contest late and operating under a strict 30-hour per-participant compute budget, this token-efficient paradigm successfully instilled robust search capabilities directly into the model's parametric memory. Our approach yielded $> \! 96\%$ validation accuracy on the bit manipulation category. Ultimately, this framework provides a practical methodology for embedding fault-tolerant search behaviors into LLMs, earning our team the \textbf{Best Fine-tuning Method} and \textbf{Best Data/Synthetic Data Method} awards (alongside a 7th Place overall finish on leaderboard) in the NVIDIA Nemotron Challenge.

\clearpage
\appendix

\titleformat{\section}{\Large\bfseries\color{darkblue}}{Appendix \thesection:}{0.5em}{}

\section{Full Chain-of-Thought (CoT) Traces}

This appendix provides two complete, unedited examples of the synthetic Chain-of-Thought (CoT) traces used to train the model. To clearly differentiate the context, the puzzle prompt is highlighted in blue, followed immediately by the model's generated reasoning trace in gray. They demonstrate the full end-to-end pipeline: from row deconstruction and minimal bitflip extraction, to the exhaustive frequency-guided Depth-First Search (DFS) with global collision verification, and finally the truth table synthesis.

\vspace{0.5cm}

\newtcblisting{promptlisting}[1]{
    listing only, 
    title=\textbf{#1}, 
    colback=blue!4, 
    colframe=darkblue!80, 
    coltitle=white,
    breakable, 
    arc=2pt,
    boxrule=0.7pt,
    left=5pt, right=5pt, top=4pt, bottom=4pt,
    listing options={basicstyle=\scriptsize\ttfamily, breaklines=true}
}

\newtcblisting{cotlisting}[1]{
    listing only, 
    title=\textbf{#1}, 
    colback=gray!5, 
    colframe=gray!70!black, 
    coltitle=white,
    breakable, 
    arc=2pt,
    boxrule=0.7pt,
    left=5pt, right=5pt, top=4pt, bottom=4pt,
    listing options={basicstyle=\scriptsize\ttfamily, breaklines=true}
}

\begin{promptlisting}{Example Trace 1: Prompt (ID: c200810b)}
In Alice's Wonderland, a secret bit manipulation rule transforms 8-bit binary numbers. The transformation involves operations like bit shifts, rotations, XOR, AND, OR, NOT, and possibly majority or choice functions.

Here are some examples of input -> output:
11001101 -> 00000000
10010111 -> 00000000
11000111 -> 00000000
11010111 -> 00000010
11110011 -> 00000110
11101111 -> 00000101
10100100 -> 00000100

Now, determine the output for: 01000101

EXPECTED ANSWER: 00000000
\end{promptlisting}
\vspace{-0.3cm} 
\begin{cotlisting}{Model Thoughts (Reasoning Trace)}
**Notation:**
- `E`: Example. `E1.7` means Example 1, Bit 7 (MSB).
- `x`: Original input bit.
- `R`, `C`, `L`: Right shift, Circular left rotation, Left shift.
- `R1` means input shifted right by 1, `C2` rotated left by 2, etc.
Format: E[Ex].[Bit]: x R[1-7] C[1-7] L[1-7]

Examples:
1. 11001101 -> 00000000
2. 10010111 -> 00000000
3. 11000111 -> 00000000
4. 11010111 -> 00000010
5. 11110011 -> 00000110
6. 11101111 -> 00000101
7. 10100100 -> 00000100

Row Creation Trace:
Input: 11001101 -> Target Output: 00000000
Bit 7 (MSB): Original bit x=1. Target O=0.
R[1-7]: Shift input right by 1-7, extract Bit 7. Result: 0000000
C[1-7]: Rotate input left by 1-7, extract Bit 7. Result: 1001101
L[1-7]: Shift input left by 1-7, extract Bit 7. Result: 1001101
E1.7: 1 0000000 1001101 1001101

Output 0 Bits: E1.7 E1.6 E1.5 E1.4 E1.3 E1.2 E1.1 E1.0 E2.7 E2.6 E2.5 E2.4 E2.3 E2.2 E2.1 E2.0 E3.7 E3.6 E3.5 E3.4 E3.3 E3.2 E3.1 E3.0 E4.7 E4.6 E4.5 E4.4 E4.3 E4.2 E4.0 E5.7 E5.6 E5.5 E5.4 E5.3 E5.0 E6.7 E6.6 E6.5 E6.4 E6.3 E6.1 E7.7 E7.6 E7.5 E7.4 E7.3 E7.1 E7.0
Number of Output 0 Bits: 50
Output 1 Bits: E4.1 E5.2 E5.1 E6.2 E6.0 E7.2
Number of Output 1 Bits: 6

Output 1 (Minority):
E4.1: 1 1000000 1010110 1110101
E5.2: 0 1100000 0111100 1111110
E5.1: 1 1000000 0011110 1111100
E6.2: 1 1100000 1011100 1111101
E6.0: 1 0000000 1110111 1110111
E7.2: 1 0000000 0010100 0010100

Output 0 (Majority):
E1.7: 1 0000000 1001101 1001101
E1.6: 1 1000000 0011010 0011011
E1.5: 0 1100000 0110100 0110111
E1.4: 0 0110000 1101000 1101110
E1.3: 1 0011000 1010000 1011100
E1.2: 1 1001100 0100000 0111001
E1.1: 0 1100110 1000000 1110011
E1.0: 1 0110011 0000000 1100110
[... Extraneous rows omitted for brevity ...]

Minimal Bit Flips:
To optimize Set Cover, we map the minority class (O=1) to the closest majority class (O=0). At least one feature MUST flip to explain the output change.
E4.1 -> E2.1 [R5,C3], E3.1 [R3,C5]
E5.2 -> E4.2 [x,R3,C5], E6.3 [x,L3,R5]
E5.1 -> E1.3 [L3,R5,R6,C2], E3.2 [L2,R3,R6,C5]
E6.2 -> E4.3 [x,L3,R5]
E6.0 -> E1.0 [R1,R5,C3,C7], E3.0 [R3,R5,C3,C5]
E7.2 -> E1.3 [L1,L3,R4,R5,C1,C4], E2.4 [L2,L3,L4,R5,C2,C4], E7.7 [L2,L5,R3,R5,C2,C3], E7.5 [L3,R2,R3,R5,C5,C6]

Unique Flip Traces:
1: [R5,C3]
2: [R3,C5]
3: [x,R3,C5]
4: [x,L3,R5]
5: [L3,R5,R6,C2]
6: [L2,R3,R6,C5]
7: [R1,R5,C3,C7]
8: [R3,R5,C3,C5]
9: [L1,L3,R4,R5,C1,C4]
10: [L2,L3,L4,R5,C2,C4]
11: [L2,L5,R3,R5,C2,C3]
12: [L3,R2,R3,R5,C5,C6]

**Set Cover Deduction Algorithm:**
1. **Mandatory Bases:** Any feature in a 1-bit flip trace is an absolute constraint and is immediately locked.
2. **Greedy Branching:** For remaining traces, iteratively select the feature with the highest intersection frequency.
3. **Collision Verification & Backtracking:** When all traces are covered, generate a Truth Table across all 64 rows. If identical inputs yield different outputs (a collision), the set is invalid. We backtrack and try the next candidate.

<deduction>
Bases: {}
Uncovered: 1,2,3,4,5,6,7,8,9,10,11,12

Freqs: R5:9 R3:6 C5:5 L3:5 C3:4
Attempting: R5
Bases: {R5}
Uncovered: 2,3,6

Freqs: R3:3 C5:3 x:1 L2:1 R6:1
Attempting: R3
Bases: {R5,R3}
Uncovered: None

All traces covered. Validating against 64-row global table:

Test TT for [R5,R3]:
|R5|R3||Out|Ref|
|---|---||---|---|
|0|0||0|E1.7,E1.6,...|
|0|1||0|E1.4,E1.3,...|
|1|0||0|E1.2,E1.1,...|
|1|1||1|E4.1,E5.2,...|
No collisions. 64-row check passed.

Conclusion: Set successfully covered and verified against full data.
Final Bases: [R3,R5]
</deduction>

Final Truth Table:
|R3|R5||Out|
|---|---||---|
|0|0||0|
|0|1||0|
|1|0||0|
|1|1||1|

Target:
Input: 01000101
Bases: [R3,R5]
R3: 00001000
R5: 00000010
Eval:
Bit 7: (0,0) -> 0
Bit 6: (0,0) -> 0
Bit 5: (0,0) -> 0
Bit 4: (0,0) -> 0
Bit 3: (1,0) -> 0
Bit 2: (0,0) -> 0
Bit 1: (0,1) -> 0
Bit 0: (0,0) -> 0

-------------------- MODEL RESPONSE (Content) --------------------
\boxed{00000000}
\end{cotlisting}

\vspace{0.8cm}

\begin{promptlisting}{Example Trace 2: Prompt (ID: 4ba4a7ec)}
In Alice's Wonderland, a secret bit manipulation rule transforms 8-bit binary numbers. The transformation involves operations like bit shifts, rotations, XOR, AND, OR, NOT, and possibly majority or choice functions.

Here are some examples of input -> output:
00001101 -> 00111111
11110111 -> 10111111
10110111 -> 10111111
00110101 -> 00111111
00111101 -> 00111111
01001100 -> 11111111
10010101 -> 00111111
11001010 -> 01111111
00010001 -> 00111111
11011101 -> 00111111

Now, determine the output for: 11000010

EXPECTED ANSWER: 01111111
\end{promptlisting}
\vspace{-0.3cm} 
\begin{cotlisting}{Model Thoughts (Reasoning Trace)}
**Notation:**
- `E`: Example. `E1.7` means Example 1, Bit 7 (MSB).
- `x`: Original input bit.
- `R`, `C`, `L`: Right shift, Circular left rotation, Left shift.
- `R1` means input shifted right by 1, `C2` rotated left by 2, etc.
Format: E[Ex].[Bit]: x R[1-7] C[1-7] L[1-7]

Examples:
1. 00001101 -> 00111111
2. 11110111 -> 10111111
3. 10110111 -> 10111111
4. 00110101 -> 00111111
5. 00111101 -> 00111111
6. 01001100 -> 11111111
7. 10010101 -> 00111111
8. 11001010 -> 01111111
9. 00010001 -> 00111111
10. 11011101 -> 00111111

Row Creation Trace:
Input: 00001101 -> Target Output: 00111111
Bit 7 (MSB): Original bit x=0. Target O=0.
R[1-7]: Shift input right by 1-7, extract Bit 7. Result: 0000000
C[1-7]: Rotate input left by 1-7, extract Bit 7. Result: 0001101
L[1-7]: Shift input left by 1-7, extract Bit 7. Result: 0001101
E1.7: 0 0000000 0001101 0001101

Output 0 Bits: E1.7 E1.6 E2.6 E3.6 E4.7 E4.6 E5.7 E5.6 E7.7 E7.6 E8.7 E9.7 E9.6 E10.7 E10.6
Number of Output 0 Bits: 15
Output 1 Bits: E1.5 E1.4 E1.3 E1.2 E1.1 E1.0 E2.7 E2.5 E2.4 E2.3 E2.2 E2.1 E2.0 E3.7 E3.5 E3.4 E3.3 E3.2 E3.1 E3.0 E4.5 E4.4 E4.3 E4.2 E4.1 E4.0 E5.5 E5.4 E5.3 E5.2 E5.1 E5.0 E6.7 E6.6 E6.5 E6.4 E6.3 E6.2 E6.1 E6.0 E7.5 E7.4 E7.3 E7.2 E7.1 E7.0 E8.6 E8.5 E8.4 E8.3 E8.2 E8.1 E8.0 E9.5 E9.4 E9.3 E9.2 E9.1 E9.0 E10.5 E10.4 E10.3 E10.2 E10.1 E10.0
Number of Output 1 Bits: 65

Output 0 (Minority):
E1.7: 0 0000000 0001101 0001101
E1.6: 0 0000000 0011010 0011010
E2.6: 1 1000000 1101111 1101110
E3.6: 0 1000000 1101111 1101110
E4.7: 0 0000000 0110101 0110101
E4.6: 0 0000000 1101010 1101010
E5.7: 0 0000000 0111101 0111101
E5.6: 0 0000000 1111010 1111010
E7.7: 1 0000000 0010101 0010101
E7.6: 0 1000000 0101011 0101010
E8.7: 1 0000000 1001010 1001010
E9.7: 0 0000000 0010001 0010001
E9.6: 0 0000000 0100010 0100010
E10.7: 1 0000000 1011101 1011101
E10.6: 1 1000000 0111011 0111010

Output 1 (Majority):
E1.5: 0 0000000 0110100 0110100
E1.4: 0 0000000 1101000 1101000
E1.3: 1 0000000 1010000 1010000
E1.2: 1 1000000 0100001 0100000
E1.1: 0 1100000 1000011 1000000
E1.0: 1 0110000 0000110 0000000
E2.7: 1 0000000 1110111 1110111
E2.5: 1 1100000 1011111 1011100
E2.4: 1 1110000 0111111 0111000
E2.3: 0 1111000 1111111 1110000
E2.2: 1 0111100 1111110 1100000
E2.1: 1 1011110 1111101 1000000
E2.0: 1 1101111 1111011 0000000
E3.7: 1 0000000 0110111 0110111
E3.5: 1 0100000 1011110 1011100
E3.4: 1 1010000 0111101 0111000
E3.3: 0 1101000 1111011 1110000
E3.2: 1 0110100 1110110 1100000
E3.1: 1 1011010 1101101 1000000
E3.0: 1 1101101 1011011 0000000
E4.5: 1 0000000 1010100 1010100
E4.4: 1 1000000 0101001 0101000
E4.3: 0 1100000 1010011 1010000
E4.2: 1 0110000 0100110 0100000
E4.1: 0 1011000 1001101 1000000
E4.0: 1 0101100 0011010 0000000
E5.5: 1 0000000 1110100 1110100
E5.4: 1 1000000 1101001 1101000
E5.3: 1 1100000 1010011 1010000
E5.2: 1 1110000 0100111 0100000
E5.1: 0 1111000 1001111 1000000
E5.0: 1 0111100 0011110 0000000
E6.7: 0 0000000 1001100 1001100
E6.6: 1 0000000 0011000 0011000
E6.5: 0 1000000 0110001 0110000
E6.4: 0 0100000 1100010 1100000
E6.3: 1 0010000 1000100 1000000
E6.2: 1 1001000 0001001 0000000
E6.1: 0 1100100 0010011 0000000
E6.0: 0 0110010 0100110 0000000
E7.5: 0 0100000 1010110 1010100
E7.4: 1 0010000 0101100 0101000
E7.3: 0 1001000 1011001 1010000
E7.2: 1 0100100 0110010 0100000
E7.1: 0 1010010 1100101 1000000
E7.0: 1 0101001 1001010 0000000
E8.6: 1 1000000 0010101 0010100
E8.5: 0 1100000 0101011 0101000
E8.4: 0 0110000 1010110 1010000
E8.3: 1 0011000 0101100 0100000
E8.2: 0 1001100 1011001 1000000
E8.1: 1 0100110 0110010 0000000
E8.0: 0 1010011 1100101 0000000
E9.5: 0 0000000 1000100 1000100
E9.4: 1 0000000 0001000 0001000
E9.3: 0 1000000 0010001 0010000
E9.2: 0 0100000 0100010 0100000
E9.1: 0 0010000 1000100 1000000
E9.0: 1 0001000 0001000 0000000
E10.5: 0 1100000 1110111 1110100
E10.4: 1 0110000 1101110 1101000
E10.3: 1 1011000 1011101 1010000
E10.2: 1 1101100 0111011 0100000
E10.1: 0 1110110 1110111 1000000
E10.0: 1 0111011 1101110 0000000

Minimal Bit Flips:
To optimize Set Cover, we map the minority class (O=0) to the closest majority class (O=1). At least one feature MUST flip to explain the output change.
E1.7 -> E6.7 [C1,C7,L1,L7]
E1.6 -> E6.6 [x,C6,L6]
E2.6 -> E5.4 [C5,C6,L5,L6]
E3.6 -> E5.4 [x,C5,C6,L5,L6]
E4.7 -> E1.5 [C7,L7]
E4.6 -> E1.4 [C6,L6]
E5.7 -> E1.5 [C4,C7,L4,L7]
E5.6 -> E1.4 [C3,C6,L3,L6]
E7.7 -> E8.6 [R1,L7]
E7.6 -> E8.5 [R2,L6]
E8.7 -> E9.4 [C1,C6,L1,L6]
E9.7 -> E9.3 [R1,L7]
E9.6 -> E9.2 [R2,L6]
E10.7 -> E2.5 [R1,R2,C6,L7], E3.5 [R2,C6,C7,L7], E4.5 [C4,C7,L4,L7]
E10.6 -> E2.4 [R2,R3,C5,L6], E3.4 [R3,C5,C6,L6], E4.4 [C3,C6,L3,L6]

Unique Flip Traces:
1: [C1,C7,L1,L7]
2: [x,C6,L6]
3: [C5,C6,L5,L6]
4: [x,C5,C6,L5,L6]
5: [C7,L7]
6: [C6,L6]
7: [C4,C7,L4,L7]
8: [C3,C6,L3,L6]
9: [R1,L7]
10: [R2,L6]
11: [C1,C6,L1,L6]
12: [R1,R2,C6,L7]
13: [R2,C6,C7,L7]
14: [R2,R3,C5,L6]
15: [R3,C5,C6,L6]

**Set Cover Deduction Algorithm:**
1. **Mandatory Bases:** Any feature in a 1-bit flip trace is an absolute constraint and is immediately locked.
2. **Greedy Branching:** For remaining traces, iteratively select the feature with the highest intersection frequency.
3. **Collision Verification & Backtracking:** When all traces are covered, generate a Truth Table across all 64 rows. If identical inputs yield different outputs (a collision), the set is invalid. We backtrack and try the next candidate.

<deduction>
Bases: {}
Uncovered: 1,2,3,4,5,6,7,8,9,10,11,12,13,14,15

Freqs: C6:9 L6:9 L7:6 C7:4 C5:4
Attempting: C6
Bases: {C6}
Uncovered: 1,5,7,9,10,14

Freqs: L7:4 C7:3 R2:2 L6:2 C1:1
Attempting: L7
Bases: {C6,L7}
Uncovered: 10,14

Freqs: R2:2 L6:2 R3:1 C5:1
Attempting: R2
Bases: {C6,L7,R2}
Uncovered: None

All traces covered. Validating against 64-row global table:

Test TT for [C6,L7,R2]:
|C6|L7|R2||Out|Ref|
|---|---|---||---|---|
|0|0|0||1|E1.5,E1.4,...|
|0|1|0||0|E1.7,E4.7,...|
|1|0|0||0|E1.6,E2.6,...|
|1|0|1||1|E1.1,E1.0,...|
|1|1|0||1|E2.7,E3.7|
No collisions, but sub-optimal to target bounds. Backtracking... (BT #1)
Removing R2. Backtrack #2.

Freqs: R2:2 L6:2 R3:1 C5:1
Attempting: L6
Bases: {C6,L7,L6}
Uncovered: None

All traces covered. Validating against 64-row global table:

Test TT for [C6,L7,L6]:
|C6|L7|L6||Out|Ref|
|---|---|---||---|---|
|0|0|0||1|E1.5,E1.4,...|
|0|1|0||0|E1.7,E4.7,...|
|1|0|0||1|E1.1,E1.0,...|
|1|0|1||0|E1.6,E2.6,...|
|1|1|1||1|E2.7,E3.7|
No collisions, but sub-optimal to target bounds. Backtracking... (BT #3)
Removing L6. Backtrack #4.
Removing L7. Backtrack #5.

Freqs: L7:4 C7:3 R2:2 L6:2 C1:1
Attempting: C7
Bases: {C6,C7}
Uncovered: 9,10,14

Freqs: R2:2 L6:2 R1:1 L7:1 R3:1
Attempting: R2
Bases: {C6,C7,R2}
Uncovered: 9

Limit K=3 reached without covering all traces. Backtracking...
Removing R2. Backtrack #6.

Freqs: R2:2 L6:2 R1:1 L7:1 R3:1
Attempting: L6
Bases: {C6,C7,L6}
Uncovered: 9

Limit K=3 reached without covering all traces. Backtracking...
Removing L6. Backtrack #7.

Freqs: R2:2 L6:2 R1:1 L7:1 R3:1
Attempting: L7
Bases: {C6,C7,L7}
Uncovered: 10,14

Limit K=3 reached without covering all traces. Backtracking...
Removing L7. Backtrack #8.
Removing C7. Backtrack #9.

Freqs: L7:4 C7:3 R2:2 L6:2 C1:1
Attempting: L6
Bases: {C6,L6}
Uncovered: 1,5,7,9

Freqs: L7:4 C7:3 C1:1 L1:1 C4:1
Attempting: L7
Bases: {C6,L6,L7}
Uncovered: None

All traces covered. Validating against 64-row global table:

Test TT for [C6,L6,L7]:
|C6|L6|L7||Out|Ref|
|---|---|---||---|---|
|0|0|0||1|E1.5,E1.4,...|
|0|0|1||0|E1.7,E4.7,...|
|1|0|0||1|E1.1,E1.0,...|
|1|1|0||0|E1.6,E2.6,...|
|1|1|1||1|E2.7,E3.7|
No collisions, but sub-optimal to target bounds. Backtracking... (BT #10)
Removing L7. Backtrack #11.

Freqs: L7:4 C7:3 C1:1 L1:1 C4:1
Attempting: C7
Bases: {C6,L6,C7}
Uncovered: 9

Limit K=3 reached without covering all traces. Backtracking...
Removing C7. Backtrack #12.
Removing L6. Backtrack #13.
Removing C6. Backtrack #14.

Freqs: C6:9 L6:9 L7:6 C7:4 C5:4
Attempting: L6
Bases: {L6}
Uncovered: 1,5,7,9,12,13

Freqs: L7:6 C7:4 R1:2 R2:2 C6:2
Attempting: L7
Bases: {L6,L7}
Uncovered: None

All traces covered. Validating against 64-row global table:

Test TT for [L6,L7]:
|L6|L7||Out|Ref|
|---|---||---|---|
|0|0||1|E1.5,E1.4,...|
|0|1||0|E1.7,E4.7,...|
|1|0||0|E1.6,E2.6,...|
|1|1||1|E2.7,E3.7|
No collisions. 64-row check passed.

Conclusion: Set successfully covered and verified against full data.
Final Bases: [L6,L7]
</deduction>

Final Truth Table:
|L6|L7||Out|
|---|---||---|
|0|0||1|
|0|1||0|
|1|0||0|
|1|1||1|

Target:
Input: 11000010
Bases: [L6,L7]
L6: 10000000
L7: 00000000
Eval:
Bit 7: (1,0) -> 0
Bit 6: (0,0) -> 1
Bit 5: (0,0) -> 1
Bit 4: (0,0) -> 1
Bit 3: (0,0) -> 1
Bit 2: (0,0) -> 1
Bit 1: (0,0) -> 1
Bit 0: (0,0) -> 1

<output>
\boxed{01111111}
</output>
\end{cotlisting}

\end{document}